\documentclass{article}
\usepackage{iclr2027_conference,times}

\usepackage{amsmath,amsfonts,bm}

\def\eqref#1{equation~\ref{#1}}

\def\1{\bm{1}}

\DeclareMathAlphabet{\mathsfit}{\encodingdefault}{\sfdefault}{m}{sl}
\SetMathAlphabet{\mathsfit}{bold}{\encodingdefault}{\sfdefault}{bx}{n}

\usepackage[utf8]{inputenc}
\usepackage[T1]{fontenc}
\usepackage[table,dvipsnames]{xcolor}
\usepackage{hyperref}
\usepackage{url}
\usepackage{booktabs}
\usepackage{amsfonts}
\usepackage{microtype}
\usepackage{adjustbox}
\usepackage{tcolorbox}
\tcbuselibrary{most}
\usepackage{etoc}
\usepackage{enumitem}
\usepackage{amsmath}
\usepackage{amssymb}
\usepackage{graphicx}
\usepackage{multirow}
\usepackage{wrapfig}
\usepackage{threeparttable}
\usepackage{makecell}
\usepackage{calligra}

\definecolor{junoorange}{RGB}{190,75,25}

\title{{\color{junoorange}\calligra Juno }: Taming Predictive Latents\\ for Vision-Language-Action Models}

\iclrfinalcopy

\author{
Yuchen Zhu$^{1,2}$ \quad
Chenyi Xu$^{2,3}$ \quad
Yulin Zhang$^{4}$ \quad
Gang Xu$^{2}$ \quad
Wentao Zhu$^{2}$\\[2pt]
$^{1}$University of Science and Technology of China \quad
$^{2}$Eastern Institute of Technology, Ningbo \\
$^{3}$Hangzhou Dianzi University \quad
$^{4}$ShanghaiTech University\\[8pt]
\normalfont\small
\textcolor{black!70}{~~Project page}%
\hspace{0.42em}%
\textcolor{junoorange}{\textperiodcentered}%
\hspace{0.42em}%
\begingroup\hypersetup{hidelinks}%
\href{https://juno-policy.github.io/}{\textcolor{junoorange}{https://juno-policy.github.io/}}%
\endgroup
}

\newcommand{\methodname}{Juno}
\begin{document}
\etocdepthtag.toc{main}

\maketitle
\fancyhead[L]{}
\renewcommand{\headrulewidth}{0pt}
\vspace{-15pt}
\begin{abstract}
Joint-embedding predictive architectures (JEPAs) predict masked or future
observations in representation space, offering a natural source of predictive
latents for vision--language--action (VLA) models. Yet making these latents
useful across pretraining, policy learning, and deployment requires addressing
three failures: mismatch with embodiment-specific control, interference with
action learning, and teacher miscalibration under distribution shifts. We
introduce {\color{junoorange}\textbf{ \textit{Juno}}}, a unified framework built around one
action-conditioned JEPA that serves as a control-aligned representation
backbone, a predictive teacher, and an adaptable dynamics model. During
\textit{pretraining}, we train it on embodiment-matched trajectories and use a dynamic
CLS loss to transfer motion-weighted patch dynamics to a compact global state.
During \textit{policy learning}, we fuse current-frame JEPA patches into VLA perception
and use a decoupled reasoning branch with separate transformation parameters
to distill future latent states for action generation. During \textit{deployment}, we
adapt the world model on all observed transitions, including failed rollouts,
freeze the adapted teacher, and re-align the policy on verified executions
using LoRA adapters and a trainable action head, without expert corrections or
task rewards. On SimplerEnv, \methodname{} raises average success from
$60.9\%$ to $68.5\%$ over Qwen3GR00T, the strongest baseline, and test-time
adaptation further reaches $72.7\%$; on a real robot, it retains
$70\%$--$75\%$ success under background, height, and object shifts where the
base policy collapses to $0\%$.
\end{abstract}

\begin{figure}[h]
    \vspace{-10pt}
    \centering
    \includegraphics[width=0.9\linewidth]{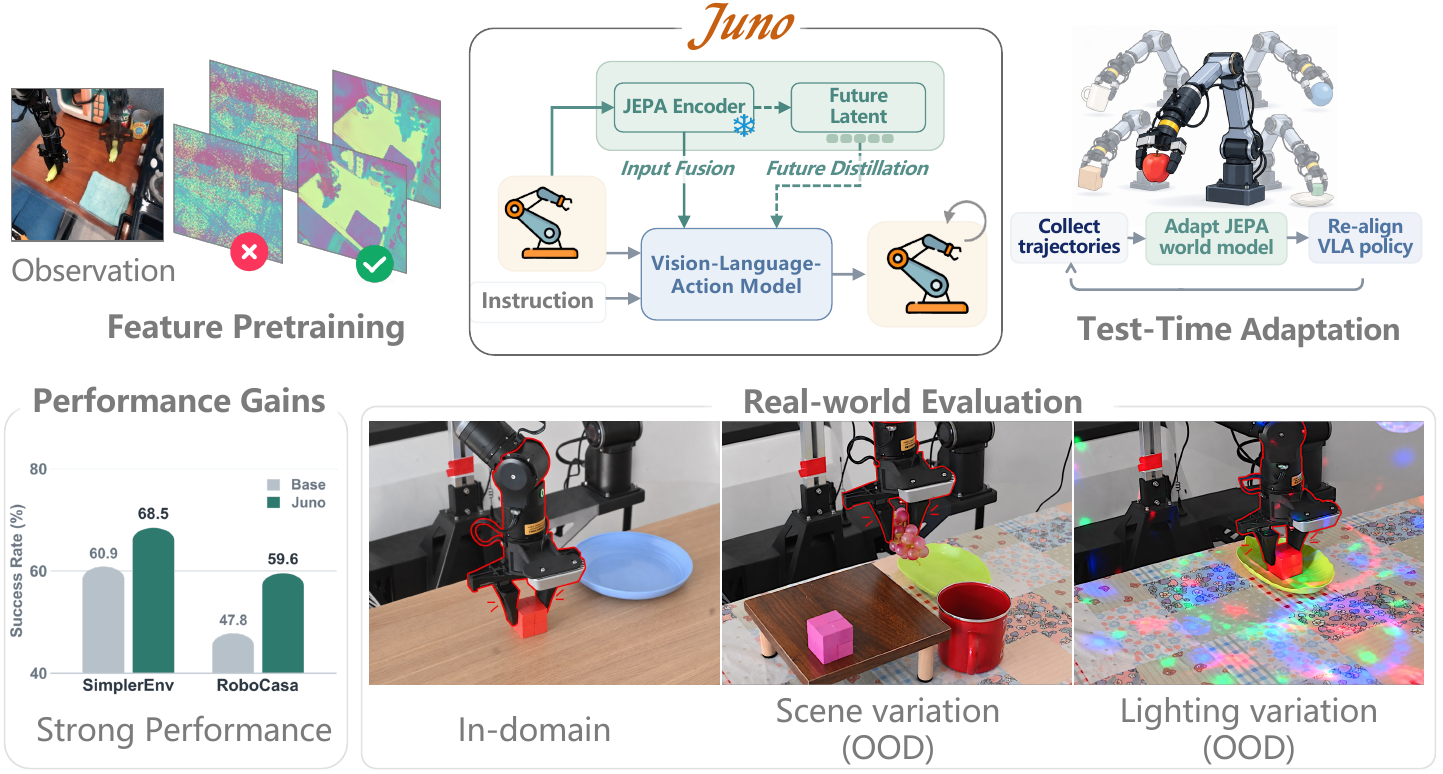}
    \caption{We present Juno, a unified framework that makes predictive latents useful across the vision-language-action (VLA) lifecycle. Juno integrates control-aligned JEPA pretraining, visual feature fusion and decoupled future-latent distillation, and dynamics-first test-time adaptation that updates the world model before re-aligning the policy. It improves performance on SimplerEnv and RoboCasa and supports robust real-world manipulation under distribution shifts.} 
    \label{fig:teaser}
\end{figure}
\section{Introduction}
Joint-embedding predictive architectures (JEPAs) predict masked or future
observations in representation space rather than in
pixels~\citep{lecun2022path,baevski2022data2vec,assran2023self,bardes2024revisiting}.
This principle has progressed from image and video level self-supervision
to action-conditioned prediction for embodied
agents~\citep{assran2025v,maes2026leworldmodel}. A growing number of
vision--language--action (VLA) systems further show that such predictive
latents benefit policy learning, whether through latent world models~\citep{sun2026vla,wang2026repwam},
future-representation alignment~\citep{zheng2025flare,zhao2026frappe,
xu2026futurevla,xia2026vlaflow}, foresight tokens~\citep{ma2026internvla,qian2025physical}, or latent
subgoals~\citep{chen2026lawam,huang2026h}. The question is therefore no longer
\emph{whether} prediction helps, but \emph{how} to make it useful across
the VLA lifecycle.

Making a predictive latent useful meets three obstacles, one per stage.
At \textbf{pretraining}, representations learned from broad internet video
are visually strong but poorly matched to embodiment-specific controllable
motion~\citep{assran2025v}. At \textbf{policy learning}, predictive
objectives are typically grafted onto the shared vision--language
pathway~\citep{zheng2025flare,sun2026vla,xu2026futurevla}, where
prediction gradients interfere with control and future-aligned states need
not be action-decodable. At \textbf{deployment}, adaptation methods assume
a reliable frozen
teacher~\citep{zhang2026ttt,park2026test,yuan2026adaworldpolicy}, yet
severe shifts miscalibrate the teacher itself, and the failed rollouts
dominating deployment data cannot serve as policy targets.  We show that a
\emph{single} action-conditioned JEPA addresses all three, serving as a
control-aligned representation backbone, a predictive teacher, and an
adaptable dynamics model.  \methodname{} realizes these roles through
\textbf{Control-Aligned Pretraining},
\textbf{Decoupled Predictive Reasoning}, and
\textbf{Dynamics-First Test-Time Adaptation}.

For representation learning, we choose a lightweight action-conditioned
JEPA~\citep{maes2026leworldmodel} and train it from pixels on the same
embodiment-matched trajectories used for policy learning. This choice is
motivated by the qualitative evidence in
Fig.~\ref{fig:method-representation}: its patch features respond locally
to robot--object interaction, while features from a large internet-video
teacher spread across the scene. Its compact global token, however,
dilutes exactly these local changes, so we introduce a dynamic CLS loss
that transfers motion-weighted patch dynamics onto the global token,
applied in a second stage with a frozen patch supervisor.

For policy learning, we fuse current-frame JEPA patch features into the
VLA's visual tokens~\citep{miao2026jepa,li2023blip} and append learnable reasoning
queries processed by a decoupled branch with separate transformation
parameters under shared causal
attention~\citep{liang2025mot,luo2026being,qian2025liquid}. The branch is distilled
against the frozen teacher's future latent states, one reasoning query
per future step, and its hidden states condition the flow-matching action
decoder. Future observations provide targets only during training; at
deployment the branch acts on the current observation and instruction.
Since the branch receives both alignment and action gradients, the aligned
state is trained to be decodable into actions rather than merely
predictive.

For deployment, even a control-aligned teacher miscalibrates under severe
shifts, and updating a policy against a miscalibrated teacher confounds
world-model error with policy error. Our key observation is an asymmetry:
failed actions are poor policy targets, but their consequences remain
valid dynamics supervision. Juno-TTT therefore adapts the world model on
all observed transitions, freezes the adapted teacher, and re-aligns the 
policy on verified executions with LoRA adapters and a trainable action head, without expert corrections or task rewards.

Our contributions follow the three stages:
\begin{itemize}[leftmargin=*, nosep]
    \item \textbf{Control-Aligned Pretraining.} We propose a dynamic CLS
    loss that aligns the global state token with motion-weighted patch
    dynamics, and show that matching the predictive teacher to the
    embodiment and control distribution can matter more than pretraining
    scale alone.
    \item \textbf{Decoupled Predictive Reasoning.} We integrate JEPA
    features into a VLA through patch-level fusion and a decoupled
    reasoning branch distilled from future latent states, so that
    predictive computation supports action generation without
    restructuring the VLA backbone.
    \item \textbf{Dynamics-First Test-Time Adaptation.} We propose a
    failure-aware adaptation procedure that exploits all deployment
    transitions, successful or not, for world-model adaptation, followed
    by conservative policy re-alignment on verified data with the adapted
    teacher held fixed.
\end{itemize}

\section{Related Work}

\textbf{Latent Predictive World Models.}
Generative world models often expend capacity on task-irrelevant visual
details \citep{hafner2025mastering,hansen2024td,zhang2020learning}, whereas
Joint-Embedding Predictive Architectures (JEPAs) predict future
representations directly, without pixel reconstruction
\citep{lecun2022path,baevski2022data2vec,assran2023self,
bardes2024revisiting,balestriero2025lejepa}. This principle now extends to
action-conditioned prediction and embodied planning
\citep{assran2025v,mur2026v,zhou2024dino,terver2025drives,zhang2026delta,
maes2026leworldmodel,sun2026vla,xu2026futurevla}, and to
latent-action abstractions from unlabeled videos
\citep{bruce2024genie,ye2025latent,bu2025univla,tharwat2025latent}. Yet none of
these works determines which predictive state best matches a particular
robot controller.

\textbf{World Action Models.}
VLA policies increasingly incorporate world-model signals for temporal
reasoning
\citep{zheng2025flare,sun2026vla,chen2026lawam,wu2024unleashing,cen2025worldvla}:
representation-centric methods align policy states with future-observation
embeddings from visual teachers
\citep{miao2026jepa,zhao2026frappe,xu2026futurevla,xia2026vlaflow}, while
others expose predicted dynamics to action generation through foresight
tokens, latent subgoals, or aligned visual--action tokens
\citep{ma2026internvla,huang2026h,su2026world,hu2024video,
wang2026repwam,luo2026being}. 
Across these granularities, however, future
alignment alone does not show that the aligned policy state can be decoded into the corresponding action. 

\textbf{Test-Time Training in Robotics.}
Test-time training (TTT) adapts models to deployment inputs via
self-supervised inner-loop optimization, without revisiting the training
set
\citep{sun2020test,wang2020tent,wang2022continual,hansen2020self,liu2021ttt++}.
In robotics, test-time updates either adapt latent prompts, adapters, or
policy components via state-grounding, future-consistency,
prediction-error, or progress signals
\citep{zhang2026ttt,park2026test,yuan2026adaworldpolicy,bai2025evolve,ji2026vane},
or serve as adaptive memory over interaction histories and unlabeled videos
\citep{sun2024learning,zhang2026test,jiang2026robottt,feng2026wamtttsteeringworldactionmodels, wang2026adajepa}.
These methods, however, all assume a reliable frozen teacher: under a
severe shift, the world model that provides the adaptation target may
itself be miscalibrated.

\section{Method}

\begin{figure}[t]
    \centering
    \includegraphics[width=0.99\linewidth]{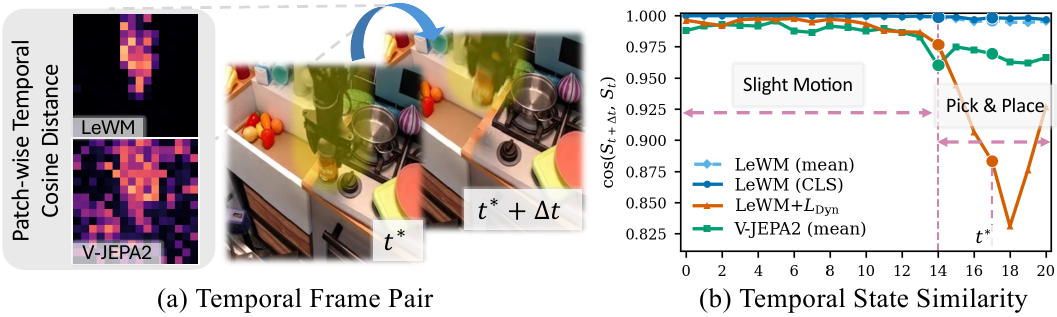}
    \caption{\textbf{(a)}~Patch-wise temporal cosine distance between frames
    $t^{\star}$ and $t^{\star}{+}\Delta t$ for LeWM and V-JEPA2.
    \textbf{(b)}~Temporal state similarity $\cos(c_t,c_{t+\Delta t})$ over the
    episode, spanning slight motion ($t\le 14$) and pick-and-place ($t>14$).
    LeWM (CLS) stays nearly constant, whereas LeWM+$L_{\mathrm{dyn}}$ becomes
    sensitive during pick-and-place; mean-pooled baselines are shown for reference.}
    \vspace{-15pt}
    \label{fig:method-representation}
\end{figure}

\subsection{Preliminaries}
\label{sec:preliminaries}
\noindent \textbf{Vision-Language-Action Policy.} We consider a robot trajectory
$\tau=\{(o_t,a_t,o_{t+1})\}_{t=1}^{T}$, where $o_t$ is the current
visual observation and $a_t$ is the executed action (or an action chunk).
We write $\mathcal{D}$ for the corresponding transition distribution.
A VLA policy maps an observation and a language
instruction $\ell$ (together with proprioceptive state when available) to an
action distribution $\pi_{\psi}(a_t\mid o_t,\ell)$ parameterized by
$\psi$.

\noindent \textbf{JEPA World Model.} A joint-embedding predictive architecture
(JEPA) encodes an observation into a latent state and predicts the latent state
of a future observation rather than reconstructing pixels~\citep{lecun2022path,assran2023self}.
Following LeWM~\citep{maes2026leworldmodel}, we use an action-conditioned JEPA
world model with an encoder $f_{\theta}$ and a latent predictor $g_{\phi}$:
\begin{equation}
    [c_t,P_t]=f_{\theta}(o_t),\qquad
    \hat c_{t+1}=g_{\phi}(c_t,a_t),
    \qquad
    \mathcal{L}_{\mathrm{JEPA}}
    =\mathcal{L}_{\mathrm{pred}}(\hat c_{t+1},c_{t+1})
      +\lambda_{\mathrm{reg}}\mathcal{R}(c),
    \label{eq:prelim-jepa}
\end{equation}
where $c_t\in\mathbb{R}^{D}$ denotes the CLS token,
$P_t=[p_{t,1},\ldots,p_{t,M}]\in\mathbb{R}^{M\times D}$ denotes the
patch-token grid, and $\mathcal{R}$ is the latent regularizer.  The predicted
$\hat c_{t+1}$ is obtained from $(c_t,a_t)$, while $c_{t+1}$ is encoded from the
observed next frame.  Thus, $c_t$ summarizes the current observation for
latent-state prediction, and $P_t$ retains the spatial evidence used by our
Dynamic CLS objective.

\begin{figure}[t]
    \centering
    \includegraphics[width=0.99\linewidth]{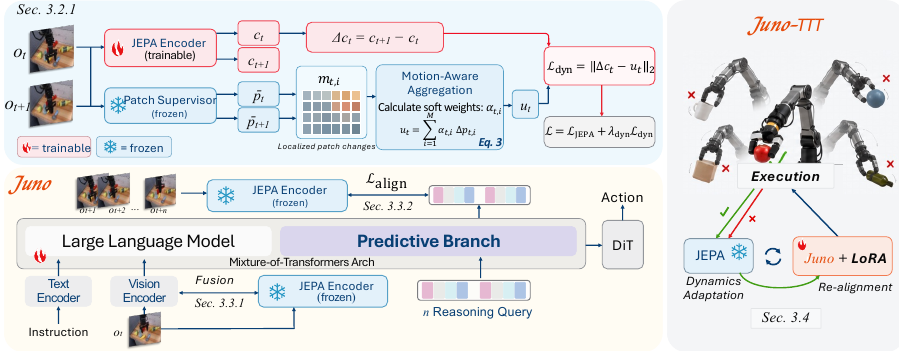}
    \caption{\textbf{Overview of \methodname.}
    \textbf{Top left:} Dynamic CLS pretraining aligns CLS transitions with motion-weighted patch changes from a frozen JEPA supervisor.
    \textbf{Bottom left:} JEPA features enrich visual tokens, while a decoupled predictive branch distills future latent states to jointly guide action generation.
    \textbf{Right:} Test-time adaptation updates the world model using both successful and failed executions, then re-aligns the policy on verified successful executions.
    Flame and snowflake icons indicate trainable and frozen modules, respectively.}
    \label{fig:method-loss}
\end{figure}

\subsection{Control-Aligned Pretraining}
\label{sec:control-aligned-representation}
We adopt LeWM as our representation backbone, a lightweight end-to-end
action-conditioned world model trained directly from pixels with roughly $15$M
parameters in its ViT-Tiny configuration~\citep{maes2026leworldmodel}.
The world model is trained from pixels on the same embodiment-matched
trajectory data used for policy learning (Sec.~\ref{sec:exp-setup}), so that
its representation is aligned with the downstream control distribution.
As shown in Fig.~\ref{fig:method-representation}(a), its patch features respond
more locally to robot--object interactions than V-JEPA 2~\citep{assran2025v}.
At the same time, its action-conditioned training objective
naturally supports test-time adaptation (Sec.~\ref{sec:ttt}).

However, to maximize planning efficiency, LeWM uses only the CLS token $c_t$
as the state for prediction and planning~\citep{lecun2022path}, while discarding the patch grid
$P_t$.  This creates a clear
bottleneck: in Fig.~\ref{fig:method-representation}(b), the CLS token changes
little during pick-and-place even though the patch tokens respond to the local
interaction.  Since the CLS token summarizes the whole scene, a sparse object
or contact change can be diluted by static background and appearance.  We
therefore use the patch-level change to supervise the CLS token, so that the
state used by the VLA remains compact while retaining control-relevant motion.

\subsubsection{Dynamic CLS Loss}
\label{sec:dynamic-cls}

As shown in Fig.~\ref{fig:method-representation}, the patch grid $P_t$
captures localized changes more clearly than the CLS token $c_t$.  We
therefore introduce an auxiliary loss that transfers patch-level changes to
the CLS representation, trained in two stages.  We first train LeWM with the
standard objective $\mathcal{L}_{\mathrm{JEPA}}$
(Eq.~\ref{eq:prelim-jepa}), whose patch tokens already respond reliably to
local interaction.  We then freeze this encoder as a patch supervisor
$\bar f$, writing $[\,\bar c_t,\bar P_t\,]=\bar f(o_t)$, and continue
training the world model under the auxiliary loss below.  The frozen
supervisor keeps the patch target stable while the trainable encoder
$f_\theta$ is being reshaped, serving as a stop-gradient reference that
cannot drift.

For each transition sampled from $\mathcal{D}$, we define the change of the
$i$-th frozen patch token as
\begin{equation}
    \Delta \bar p_{t,i}=\bar p_{t+1,i}-\bar p_{t,i},
    \qquad
    m_{t,i}=\frac{\lVert\Delta \bar p_{t,i}\rVert_2}{\sqrt{D}}.
    \label{eq:method-patch-motion}
\end{equation}
Then we do a \textit{Motion-Aware Aggregation} by converting the relative patch-change magnitudes into soft weights $\alpha_{t,i}$(Fig. \ref{fig:method-loss} Top left):
\begin{equation}
    \widetilde m_{t,i}=
    \frac{m_{t,i}}
    {\max\!\left(M^{-1}\sum_{j=1}^{M}m_{t,j},\epsilon\right)},
    \qquad
    \alpha_{t,i}=
    \frac{\exp(\widetilde m_{t,i}/\tau)}
    {\sum_{j=1}^{M}\exp(\widetilde m_{t,j}/\tau)},
    \label{eq:method-motion-weights}
\end{equation}
where $\tau$ denotes the temperature, $\epsilon$ is a small constant that
prevents division by zero, and $M$ is the number of patches.  We define the
motion-weighted patch target as
$u_t=\sum_{i=1}^{M}\alpha_{t,i}\Delta \bar p_{t,i}$ and, for the
\emph{trainable} encoder, the corresponding CLS transition as
$\Delta c_t=c_{t+1}-c_t$.
We then optimize the loss function as shown in Fig.~\ref{fig:method-loss}:
\begin{equation}
    \mathcal{L}_{\mathrm{dyn}}(\theta)=
    \mathbb{E}_{(o_t,a_t,o_{t+1})\sim\mathcal{D}}
    \left[
        \left\|\Delta c_t-u_t\right\|_2^2
    \right].
    \label{eq:method-dynamic-loss}
\end{equation}
The action $a_t$ is not directly used in this auxiliary target; it remains part
of the action-conditioned JEPA objective in Eq.~\ref{eq:prelim-jepa}.  The
second-stage objective is
\begin{equation}
    \mathcal{L}_{\mathrm{LeWM+Dyn}}
    =
    \mathcal{L}_{\mathrm{JEPA}}
    +\lambda_{\mathrm{dyn}}\mathcal{L}_{\mathrm{dyn}}.
    \label{eq:method-full-pretrain}
\end{equation}
This preserves LeWM's single-CLS interface while making $c_t$ more responsive
to localized changes captured by $P_t$.

\subsection{Decoupled Predictive Reasoning}

\subsubsection{Patch-Level Fusion}

As shown in Fig.\ref{fig:method-loss} bottom left, after pretraining, we freeze the JEPA encoder $f_\theta$ and denote it by
$E_J$.  Its current-frame patch features are projected to the VLA hidden
width by a trainable two-layer MLP and fused into the native visual tokens
through cross-attention with a gated residual connection (fixed gain
$\gamma$), which preserves the spatial information in $P_t$ while leaving
the original VLA token sequence unchanged.

\subsubsection{Future Latent-State Distillation and Action Generation}
We instantiate the predictive branch with the Mixture-of-Transformers
(MoT) decoder block \citep{liang2025mot,luo2026being}, the
vision language and predictive reasoning branches keep separate non-embedding
parameters while sharing one causal attention graph.  The vision--language
branch thus preserves the representation used for action generation, while
the reasoning branch is dedicated to predicting future JEPA states.

We append $K$ learnable reasoning queries to the vision--language sequence.
Let $H_{R,t}^{L}$ denote the final reasoning states, and let the trainable
projector $g_r$ produce the predicted future states
$\widehat Z_t=g_r(H_{R,t}^{L})$. During training, the frozen JEPA encoder
provides detached CLS targets from the $K$ future observations:
$Z_t^{+}=E_J^{\mathrm{cls}}
(o_{t+\Delta:t+K\Delta})$. Since the number of future targets matches
the number of reasoning queries, we align them one-to-one:
\begin{equation}
 \mathcal{L}_{\mathrm{align}}=
 \frac{1}{BK}\sum_{b=1}^{B}\sum_{k=1}^{K}
 \left[
 1-\operatorname{cos}(\widehat z_{b,k},z^+_{b,k})
 +\lambda_{\mathrm{sl1}}\operatorname{SmoothL1}_{\delta}
 (\widehat z_{b,k},z^+_{b,k})
 \right],
 \label{eq:method-align-loss}
\end{equation}
The full hidden sequence, including the vision--language and reasoning states,
conditions the flow-matching action decoder.  Thus, the reasoning branch
receives both future-state alignment and action-generation gradients; this
joint pathway is what makes the aligned state decodable into actions rather
than merely predictive of future observations.  The
joint objective is
\begin{equation}
 \mathcal{L}_{\mathrm{VLA}}
 =\mathcal{L}_{\mathrm{FM}}+\alpha\,\mathcal{L}_{\mathrm{align}},
 \label{eq:method-joint-loss}
\end{equation}
At inference time, the future-state target path is
removed, and the
predictive branch operates only on the current observation and language
instruction.

\subsection{Dynamics-First Test-Time Adaptation}
\label{sec:ttt}

The control-aligned pretraining and decoupled predictive reasoning above
already yield strong frozen-teacher policies in simulation and on real robots
(Sec.~\ref{sec:exp-sim}, Sec.~\ref{sec:exp-real}), and remain robust to
moderate shifts in background, layout, and object instance. Under more severe
shifts, however---strong observation noise or abrupt illumination change---the
frozen JEPA teacher itself becomes unreliable, and the policy coupled to it
collapses. We therefore add a \emph{dynamics-first} test-time adaptation stage,
motivated by a single observation:

\begin{tcolorbox}[
  enhanced,
  colback=cyan!3!white,
  colframe=blue!20!black,
  boxrule=0pt,
  leftrule=2pt,
  sharp corners,
  left=6pt,
  right=6pt,
  top=6pt,
  bottom=6pt
]
\textit{A failed action cannot teach the robot what to do,
but faithfully witnesses how the world responds.}
\end{tcolorbox}

\noindent A failed rollout still faithfully records how the executed actions
change the environment, and these transitions can be exploited without expert
correction or task reward.  Our method therefore proceeds in two stages(Fig. \ref{fig:method-loss} right):
(i)~adapt the world model to the observed test-time environment using
\emph{all} interaction trajectories as action-conditioned dynamics
supervision; (ii)~re-align the policy with the adapted representation on
verified executions only.

\noindent\textbf{Test-time data curation.}  We partition the collected
deployment buffer into two disjoint subsets: transitions from verified
successful executions, $\mathcal{D}_{\mathrm{ok}}$, and transitions from
failed executions, $\mathcal{D}_{\mathrm{fail}}$, after a lightweight
cleaning procedure that removes invalid segments (details in the appendix~\ref{app:ttt}). 
We write
$\mathcal{D}_{\mathrm{ttt}}=\mathcal{D}_{\mathrm{ok}}\cup\mathcal{D}_{\mathrm{fail}}$
for the full buffer.

\noindent\textbf{Stage 1: Dynamics adaptation.}  The world model is adapted
on the full buffer $\mathcal{D}_{\mathrm{ttt}}$ with the same
action-conditioned objective used in pretraining
(Eq.~\ref{eq:prelim-jepa}), performing a full-parameter update of both
$f_{\theta}$ and $g_{\phi}$ initialized from their pretrained weights:
\begin{equation}
    \theta^{\star},\phi^{\star}
    =\arg\min_{\theta,\phi}\;
    \mathbb{E}_{(o_t,a_t,o_{t+1})\sim\mathcal{D}_{\mathrm{ttt}}}
    \Big[
        \mathcal{L}_{\mathrm{pred}}
        \big(g_{\phi}(c_t,a_t),\,c_{t+1}\big)
        +\lambda_{\mathrm{reg}}\mathcal{R}(c)
    \Big],
    \label{eq:method-ttt-world}
\end{equation}
where $[c_t,P_t]=f_{\theta}(o_t)$ as defined in
Sec.~\ref{sec:preliminaries}.  Crucially, this objective supervises only the
\emph{consequences} of the executed actions, so failed trajectories provide
the same learning signal as successful ones at this stage.

\noindent\textbf{Stage 2: Policy re-alignment.}  The adapted encoder is
then frozen, $E_J^{\mathrm{ttt}}\!\triangleq f_{\theta^{\star}}$, and
produces the future-state targets
$\widetilde Z_t^{+}=E_J^{\mathrm{ttt,cls}}(o_{t+\Delta:t+K\Delta})$ in
place of the original teacher in Eq.~\ref{eq:method-align-loss}.  The policy
is re-aligned on verified executions only, using the same composite
objective as in training (Eq.~\ref{eq:method-joint-loss}):
\begin{equation}
    \mathcal{L}_{\mathrm{ttt}}(\psi)
    =\mathbb{E}_{\mathcal{D}_{\mathrm{ok}}}
    \Big[
        \mathcal{L}_{\mathrm{FM}}(\psi)
        +\lambda_{\mathrm{ttt}}\,
        \mathcal{L}_{\mathrm{align}}(\psi;\widetilde Z^{+})
    \Big].
    \label{eq:method-ttt-policy}
\end{equation}
For stability under scarce test-time data, the policy is updated through
LoRA adapters inserted into all policy weights while the pre-trained
backbone remains frozen; the action head is fully trainable without LoRA,
and all test-time updates use a reduced learning rate.  We refer to this two-stage
procedure as \emph{Juno-TTT}.  Note the asymmetry between the stages:
dynamics adaptation consumes every transition, whereas policy re-alignment
trusts only verified executions---failed actions never become supervision
targets.  Freezing $E_J^{\mathrm{ttt}}$ keeps the
alignment target fixed throughout policy adaptation.

\section{Experiments}
\label{sec:experiments}

We evaluate \methodname{} along three axes: manipulation performance in
simulation, transfer of the observed benefits to real-robot manipulation, and
adaptation to deployment shifts through test-time training. We first evaluate
the policy without test-time updates to assess the combined effect of
JEPA-enhanced perception and predictive specialization. Adaptation results
are distinguished from this frozen-policy evaluation because they use
additional deployment interactions and optimization.

\subsection{Experimental Setup}
\label{sec:exp-setup}

\paragraph{Benchmarks.}
We evaluate SimplerEnv\citep{li24simpler} with the WidowX embodiment and RoboCasa-GR1 Tabletop
Tasks with the GR1 embodiment. For the real-world evaluation, we use the
right arm of a bimanual AgileX Cobot Magic platform under nominal and shifted conditions,
as illustrated in Fig.~\ref{fig:real-setup}. Detailed task descriptions and
shift specifications are provided later and in the supplementary material.

\begin{tcolorbox}[colback=gray!5!white,colframe=black!75!black,
title=Real-robot benchmark]
All real-robot conditions, including in-domain evaluation, test
\textbf{spatial position generalization}: tasks are executed at positions
excluded from the demonstrations. ``In-domain'' retains the nominal
appearance and object configuration under a held-out arrangement, while OOD
conditions also vary visual or physical factors.
\end{tcolorbox}

\paragraph{Models and training.}
We use Qwen3GR00T~\citep{ye2026starvla} as the base VLA and primary baseline.
The training data are selected according to the evaluated benchmark rather than
shared across benchmarks. For SimplerEnv, we train on the BridgeData V2 and
Fractal datasets. For RoboCasa-GR1, we train one multitask model jointly on
the benchmark's own GR1 Tabletop Tasks data. For the
real-robot experiments, we initialize from a policy pretrained for SimplerEnv 
and then adapt it with a small set of simple demonstrations
collected on the this platform.

\paragraph{Evaluation protocol.}
For each benchmark, we report task success rates and their unweighted mean
across tasks. RoboCasa-GR1 is evaluated for 50 episodes per task. SimplerEnv
results are averaged over 4 evaluation runs, and we report the mean success
rate of each task. For the real-robot experiments, we run 20 trials per condition
and report the success rate.

\subsection{Simulation Benchmark Results}
\label{sec:exp-sim}

\begin{table*}[ht]
\centering
\caption{Success rates (\%) of VLA models evaluated with the WidowX robot in
the SimplerEnv simulation benchmark. The best result in each column is shown
in \textbf{bold}, and the second-best distinct result is
\underline{underlined}. Tied results receive the same formatting.
Parenthetical arrows on the last two rows are the change from Qwen3GR00T.}
\label{tab:simplerenv}
\small
\renewcommand{\arraystretch}{1.15}
\setlength{\tabcolsep}{5pt}
\begin{adjustbox}{max width=\textwidth}
\begin{tabular}{@{}lccccc@{}}
\toprule
\textbf{Method}
& \textbf{Average}
& \makecell[c]{\textbf{Put Spoon} \\ \textbf{on Towel}}
& \makecell[c]{\textbf{Put Carrot} \\ \textbf{on Plate}}
& \makecell[c]{\textbf{Stack Green Block} \\ \textbf{on Yellow Block}}
& \makecell[c]{\textbf{Put Eggplant} \\ \textbf{in Yellow Basket}} \\
\midrule

OpenVLA-OFT~\citep{kim2024openvla}
& 41.8 & 34.2 & 30.0 & 30.0 & 72.5 \\

RoboVLM~\citep{li2026matters}
& 42.7 & 50.0 & 37.5 & 0.0 & 83.3 \\

Magma~\citep{yang2025magma}
& 44.8 & 37.5 & 29.2 & 20.8 & 91.7 \\

CogACT~\citep{li2024cogact}
& 51.3 & 71.7 & 50.8 & 15.0 & 67.5 \\

SpatialVLA~\citep{qu2025spatialvla}
& 34.4 & 20.8 & 20.8 & 25.0 & 70.8 \\

TraceVLA~\citep{zheng2025tracevla}
& 27.7 & 12.5 & 16.6 & 16.6 & 65.0 \\

VideoVLA~\citep{shen2026videovla}
& 53.1 & 75.0 & 20.8 & \textbf{45.8} & 70.8 \\

VLA-JEPA~\citep{sun2026vla}
& 57.3 & 75.0 & \textbf{70.8} & 12.5 & 70.8 \\

\midrule

$\pi_0$~\citep{black2024pi_0}
& 53.1 & 29.2 & 62.5 & 29.2 & 91.6 \\

$\pi_{0.5}$~\citep{intelligence2025pi_}
& 57.1 & 49.3 & 64.7 & \underline{44.7} & 69.7 \\

Isaac-GR00T-N1.6-Bridge~\citep{gr00tn1_2025}
& 57.1 & 64.5 & \underline{65.5} & 5.5 & 93.0 \\

\midrule

Qwen3GR00T (Baseline)~\citep{ye2026starvla}
& 60.9 & 72.9 & 60.4 & 14.6 & \underline{95.8} \\

\rowcolor{gray!30}
\textbf{\methodname{} (Ours)}
& \underline{68.5}\,\textcolor{green!45!black}{\scriptsize($\uparrow$7.6)}
& \underline{86.5}\,\textcolor{green!45!black}{\scriptsize($\uparrow$13.6)}
& \textnormal{59.4}\,\textcolor{red!60!black}{\scriptsize($\downarrow$1.0)}
& \textnormal{31.3}\,\textcolor{green!45!black}{\scriptsize($\uparrow$16.7)}
& \textbf{96.9}\,\textcolor{green!45!black}{\scriptsize($\uparrow$1.1)} \\

\rowcolor{gray!30}
\textbf{$+$ \textit{Juno-TTT} (Ours)}
& \textbf{72.7}\,\textcolor{green!45!black}{\scriptsize($\uparrow$11.8)}
& \textbf{94.8}\,\textcolor{green!45!black}{\scriptsize($\uparrow$21.9)}
& \textnormal{62.5}\,\textcolor{green!45!black}{\scriptsize($\uparrow$2.1)}
& \textnormal{37.5}\,\textcolor{green!45!black}{\scriptsize($\uparrow$22.9)}
& \underline{95.8} \\

\bottomrule
\end{tabular}
\end{adjustbox}
\vspace{-0.5em}
\end{table*}

\paragraph{SimplerEnv.}
Table~\ref{tab:simplerenv} shows that \methodname{} achieves $68.5\%$ average
success, the highest among methods evaluated without TTT: $+7.6$ over
Qwen3GR00T ($60.9\%$) and $+11.2$ over the strongest prior method (VLA-JEPA,
$57.3\%$), also surpassing the $\pi_0$/$\pi_{0.5}$ and GR00T-N1.6 families
($53.1\%$--$57.1\%$). The gains are broad rather than isolated: spoon
placement exceeds every prior method by at least $11.5$ points ($86.5\%$),
block stacking more than doubles the base VLA ($14.6\%$ to $31.3\%$), and
eggplant reaches a column-best $96.9\%$, while carrot stays essentially flat
($60.4\%$ to $59.4\%$); three of four tasks improve with no meaningful
regression on the fourth.

\begin{table}[t]
  \centering
  \caption{RoboCasa-GR1 Tabletop Tasks. The 24 tasks are partitioned into five
  evaluation groups. Each entry is the unweighted mean success rate (\%) within
  the corresponding group (Overall averages over all 24 tasks). Best result in
  each column is shown in \textbf{bold}.}
  \label{tab:robocasa}
  \small
  \renewcommand{\arraystretch}{1.2}
  \setlength{\tabcolsep}{4.5pt}
  \begin{tabular}{@{}lcccccc@{}}
    \toprule
    \textbf{Method}
      & \makecell[c]{\textbf{PnP + Close}}
      & \makecell[c]{\textbf{PnP From} \\ \textbf{Cutting}\textbf{board}}
      & \makecell[c]{\textbf{PnP From} \\\textbf{Place}\textbf{mat}}
      & \makecell[c]{\textbf{PnP From} \\\textbf{Plate}}
      & \makecell[c]{\textbf{PnP From} \\\textbf{Tray}}
      & \textbf{Avg.} \\
    \midrule
    GR00T-N1.6
      & 24.2 & 56.9 & 51.9 & 57.6 & \textbf{55.1} & 47.6 \\
    Qwen3PI
      & 42.3 & 46.0 & 43.5 & 44.0 & 44.0 & 43.9 \\
    Qwen3OFT
      & 43.7 & 50.4 & 41.5 & 61.0 & 49.2 & 48.8 \\
    Qwen3FAST
      & 35.0 & 50.4 & 33.5 & 45.0 & 32.0 & 39.0 \\
 Qwen3GR00T
      & 50.3 & 52.8 & 38.0 & 58.5 & 39.2 & 47.8 \\
    \rowcolor{gray!15}
    \textbf{\methodname{} (Ours)}
      & \textbf{57.3} & \textbf{64.8} & \textbf{55.5} & \textbf{68.0} & 53.6 & \textbf{59.6} \\
    \bottomrule
  \end{tabular}
  \vspace{-10pt}
\end{table}

\paragraph{RoboCasa-GR1.}
Table~\ref{tab:robocasa} evaluates a single model across all 24 tasks.
\methodname{} achieves $59.6\%$ average success, improving on Qwen3GR00T
($47.8\%$) by $11.8$ points and on the strongest listed baseline, Qwen3OFT
($48.8\%$), by $10.8$ points.  It outperforms Qwen3GR00T on $21$ of $24$
tasks and achieves the highest score on $16$, with gains up to $46.0\%
\to 80.0\%$ (bottle placement with cabinet closing) and regressions
limited to two tasks (at most $-6$ points).  Per-task results are reported
in Appendix~\ref{app:eval-robocasa}.
\subsection{Real-World Robot Evaluation}
\label{sec:exp-real}

\begin{figure}[h]
    \centering
    \includegraphics[width=0.98\linewidth]{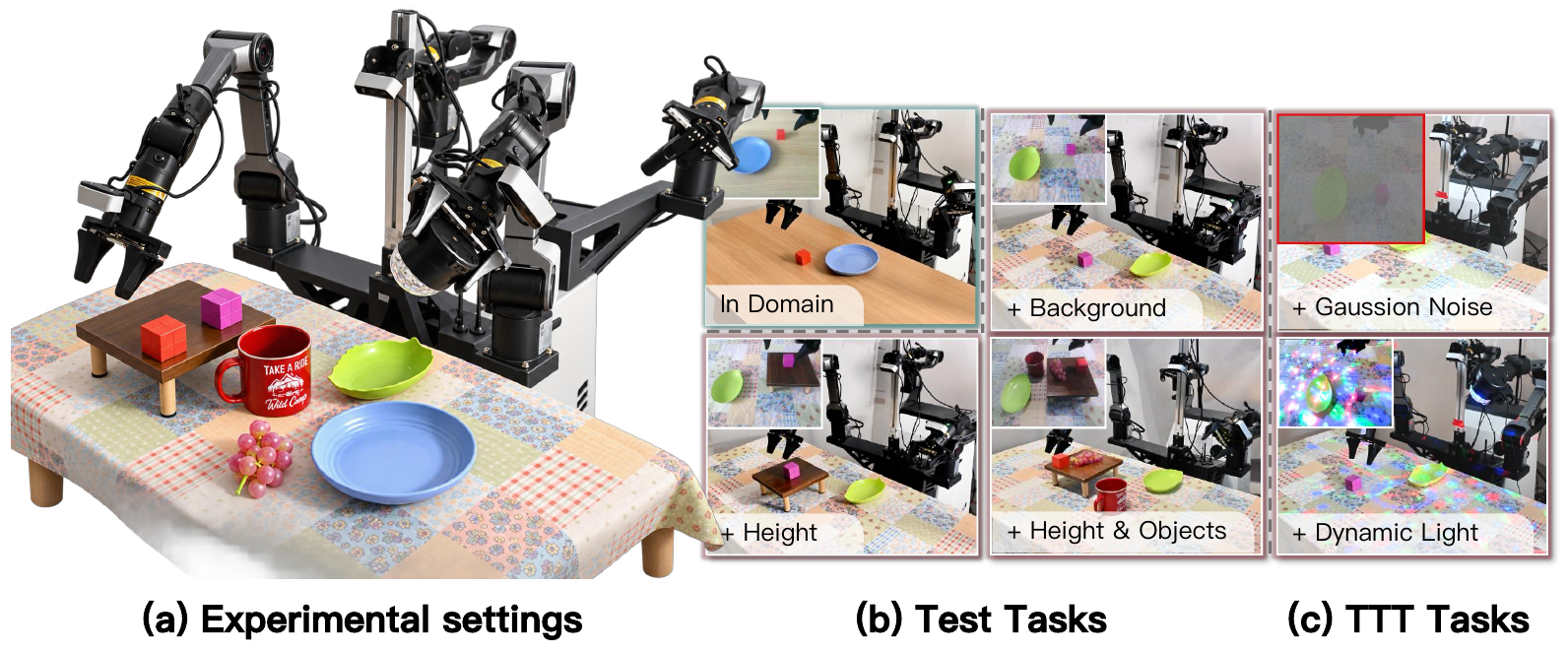}
    \caption{\textbf{Real-robot setup and evaluation conditions.}
    (a) AgileX Cobot Magic, an ALOHA-style platform; only the right PiPER arm is used.
    (b) Nominal and shifted conditions involving background, manipulation
    height, and objects for evaluating the frozen policy.
    (c) Gaussian observation noise and dynamic lighting conditions for
    test-time training, evaluated separately from the frozen-policy results.}
    \label{fig:real-setup}
    \vspace{-5pt}
\end{figure}

\begin{table}[tb]
\centering
\begin{minipage}{0.54\textwidth}
\centering
\caption{Frozen-policy real-world evaluation on the right arm of
ALOHA. Success rate (\%) over 20 trials.}
\label{tab:real-world}
\small
\renewcommand{\arraystretch}{1.25}
\setlength{\tabcolsep}{4.5pt}
\resizebox{0.96\linewidth}{!}{
    \begin{threeparttable}
    \begin{tabular}{@{}lcccc@{}}
      \toprule
      \multirow{2}{*}{\textbf{Method}}
        & \multirow{2}{*}{\textbf{In-Domain}}
        & \multicolumn{3}{c}{\textbf{Out-of-Domain}} \\
      \cmidrule(lr){3-5}
      &
        & \textbf{Background}
        & \textbf{+ Height}
        & \textbf{+ H + Object} \\
      \midrule
      Qwen3GR00T
        & 40.0\,(8/20)
        & 0.0\,(0/20)
        & 0.0\,(0/20)
        & 0.0\,(0/20) \\
      \rowcolor{gray!15}
      \textbf{\methodname{}}
        & \textbf{100.0\,(20/20)}
        & \textbf{75.0\,(15/20)}
        & \textbf{70.0\,(14/20)}
        & \textbf{70.0\,(14/20)} \\
      \bottomrule
    \end{tabular}
    \end{threeparttable}
}
\end{minipage}
\hfill
\begin{minipage}{0.44\textwidth}
\centering
\caption{Test-time training on ALOHA.
Success rate (\%) over 20 trials under Gaussian observation noise
and dynamic lighting.}
\label{tab:real-world-right}
\small
\renewcommand{\arraystretch}{1.25}
\setlength{\tabcolsep}{4.5pt}
\resizebox{0.99\linewidth}{!}{
    \begin{threeparttable}
    \begin{tabular}{@{}lcc@{}}
      \toprule
      \textbf{Method}
        & \textbf{Gaussian Noise}
        & \textbf{Dynamic Lighting} \\
      \midrule
      \textbf{\methodname{}} (frozen)
        & 40.0\,(8/20)
        & 55.0\,(11/20) \\
      \rowcolor{gray!15}
      \textbf{$+$ \textit{Juno-TTT}}
        & \textbf{65.0\,(13/20)}
        & \textbf{70.0\,(14/20)} \\
      \bottomrule
    \end{tabular}
    \end{threeparttable}
}
\end{minipage}
\vspace{-15pt}
\end{table}

\paragraph{Platform and Data Collection.}
We evaluate physical transfer on an AgileX Cobot Magic ALOHA-style
leader--follower system with 6-DoF PiPER follower arms and parallel
grippers, using only the right arm. Actions are 7-dimensional end-effector
deltas comprising translation, rotation, and a binary gripper command. After
pretraining on the base policy's public data, we collect 60 adaptation
demonstrations for \textit{``Put the red cube into the blue bowl.''}, with
20 demonstrations at each of three fixed cube positions. (More details in appendix~\ref{app:eval-real})

\paragraph{Evaluation Conditions.}
From the nominal setting, we evaluate three cumulative shifts
(Fig.~\ref{fig:real-setup}(b)): \emph{background} changes the tablecloth,
cube color, and bowl shape/color; \emph{height} additionally raises the work
plane by $10$\,cm; and \emph{object} replaces the cube with a grape and
updates the instruction. We compare \methodname{} with Qwen3GR00T over $20$
trials per condition, with all models frozen to measure generalization without
deployment adaptation.

\paragraph{In-domain manipulation performance.}
In the in-domain condition, \methodname{} succeeds in all 20 trials
($100\%$), whereas the baseline succeeds in 8 of 20 ($40\%$), a gap of
$60$ percentage points (Table~\ref{tab:real-world}). The policy therefore
solves the task reliably after adaptation from only 60 demonstrations.

\paragraph{Generalization under deployment shifts.}
The advantage widens under distribution shift. \methodname{} retains
$75\%$ success (15/20) with an out-of-distribution background and $70\%$
(14/20) under both the added height change and the combined height-plus-object
change, so performance degrades gracefully and remains stable as the shifts
accumulate. The baseline collapses to 0 of 20 trials in all three shifted
conditions, despite reaching $40\%$ in domain. Since both policies remain frozen here, these gains are attributable to the
learned representations alone; test-time adaptation brings further
improvements under more severe shifts (Sec.~\ref{sec:exp-ttt}).

\subsection{Test-Time Adaptation Results}
\label{sec:exp-ttt}
We next evaluate \textit{Juno-TTT} (Sec.~\ref{sec:ttt}) under an offline protocol:
the policy first collects a fixed set of rollouts in the deployment
environment, the two-stage adaptation runs offline, and the adapted policy
is then evaluated frozen. Collection sizes and optimization budgets are
given in Appendix~\ref{app:ttt}.

\paragraph{Simulation.}
On SimplerEnv, we apply \textit{Juno-TTT} in the nominal evaluation condition, which
isolates the benefit of deployment data in the absence of distribution
shift.  Adaptation raises average success from $68.5\%$ to $72.7\%$
(Table~\ref{tab:simplerenv}), improving three of the four tasks: spoon
placement from $86.5\%$ to $94.8\%$, block stacking from $31.3\%$ to
$37.5\%$, and carrot placement from $59.4\%$ to $62.5\%$, while eggplant
placement stays essentially unchanged ($96.9\%$ to $95.8\%$).  Thus, even
without any shift, the policy's own rollouts provide useful dynamics
supervision for further improvement.

\paragraph{Real robot.}
The benefit grows under severe deployment shifts on the PiPER arm
(Table~\ref{tab:real-world-right}), where we collect $40$ rollouts
separately in each condition and adapt the model to that condition.  Under
Gaussian observation noise, the frozen
policy retains only $40\%$ success, and \textit{Juno-TTT} recovers it to $65\%$;
under dynamic lighting, \textit{Juno-TTT} raises success from $55\%$ to $70\%$.

\subsection{Ablation Study}
\begin{wraptable}{r}{0.46\linewidth}
  \vspace{-40pt}
  \centering
  \caption{Component ablation on SimplerEnv (average success,
  \%). Each row removes one component; dark red arrows give the drop
  from the full model. Evaluation follows the main SimplerEnv protocol.}
  \label{tab:ablation}
  \small
  \renewcommand{\arraystretch}{1.12}
  \setlength{\tabcolsep}{3.5pt}
  \begin{tabular}{@{}ccccc@{}}
    \toprule
    \textbf{Fusion}
     & \textbf{MoT}
     & \textbf{$\mathcal{L}_{\mathrm{align}}$}
     & \textbf{$\mathcal{L}_{\mathrm{dyn}}$}
     & \textbf{Avg.} \\
    \midrule
               & \checkmark & \checkmark & \checkmark & 53.9\,\textcolor{red!60!black}{\scriptsize($\downarrow$14.6)} \\
    \checkmark &            & \checkmark & \checkmark & 65.4\,\textcolor{red!60!black}{\scriptsize($\downarrow$3.1)} \\
    \checkmark & \checkmark &            & \checkmark & 60.8\,\textcolor{red!60!black}{\scriptsize($\downarrow$7.7)} \\
               &            & \checkmark & \checkmark & 60.4\,\textcolor{red!60!black}{\scriptsize($\downarrow$8.1)} \\
    \checkmark & \checkmark & \checkmark &            & 65.6\,\textcolor{red!60!black}{\scriptsize($\downarrow$2.9)} \\
    \midrule
    \rowcolor{gray!30}
    \checkmark & \checkmark & \checkmark & \checkmark & \textbf{68.5} \\
    \bottomrule
  \end{tabular}
  \vspace{-8pt}
\end{wraptable}

\paragraph{Component ablation.}
Table~\ref{tab:ablation} reports a leave-one-out ablation on SimplerEnv
under the same evaluation protocol as the main results.  Fusion and the
MoT branch form a coupled pair.  Removing fusion alone causes the largest
drop ($53.9\%$), even below the base VLA, yet removing it together with
the MoT branch recovers to $60.4\%$: without the fused patch pathway,
the predictive branch imposes future-state objectives on ungrounded
representations and turns harmful.  Conversely, removing the MoT branch
alone costs only $3.1$ points ($65.4\%$), so its value is realized
through fusion.  The alignment loss is what makes this pair useful:
without $\mathcal{L}_{\mathrm{align}}$, the model gains nothing over
the base VLA ($60.8\%$ vs.\ $60.9\%$), i.e., the predictive branch
helps only when its predictions are directly supervised.  The dynamic CLS
loss contributes a smaller but consistent gain ($-2.9$).

\section{Conclusion}
We present \methodname{}, a unified framework that integrates predictive
world models into VLA pretraining, policy learning, and test-time adaptation.
By combining control-aligned representations, decoupled predictive reasoning,
and dynamics-first adaptation, \methodname{} links future-state prediction
with action generation and improves robustness through deployment experience.
Experiments in SimplerEnv, RoboCasa-GR1, and real-world manipulation show
higher success and resilience to distribution shifts. Future work
will scale up training and evaluation across larger datasets, broader tasks,
and multiple embodiments, further taming predictive latents for general-purpose
VLA control.

\clearpage
\bibliography{iclr2027_conference}
\bibliographystyle{iclr2027_conference}

\clearpage
\appendix
\renewcommand{\methodname}{\textit{Juno}}
\etocdepthtag.toc{supplementary}
\begingroup
\renewcommand{\contentsname}{Supplementary Material Contents}
\etocsettagdepth{main}{none}
\etocsettagdepth{supplementary}{subsection}
\etocsetnexttocdepth{subsection}
\tableofcontents
\endgroup
\clearpage

\section{Additional Representation Analysis}
\label{app:representation}

A useful predictive target should preserve state changes that matter for
control while remaining stable under irrelevant appearance variation.
We therefore compare two frozen encoders: V-JEPA~2, pretrained on video
\citep{assran2025v,mur2026v}, and the control-aligned encoder, denoted
\textit{Ours} in the figures. As introduced in the first part of Method,
the control-aligned encoder is trained on robot trajectories with an action-conditioned
JEPA objective and the Dynamic CLS loss
(Eq.~\ref{eq:method-dynamic-loss}).
The analyses cover spatial feature structure and the relationship between
global-token transitions and action patterns. They assess teacher
suitability in the studied control setting and do not isolate pretraining data,
architecture, and objective as independent factors.

\paragraph{Feature extraction.}
Patch-level maps use each encoder's native grid. Feature channels and
PCA colors are not assumed to correspond across encoders.

\subsection{Attention and PCA Visualization}
\label{app:attention}

\begin{figure}[t]
 \centering
 \includegraphics[width=0.95\linewidth]{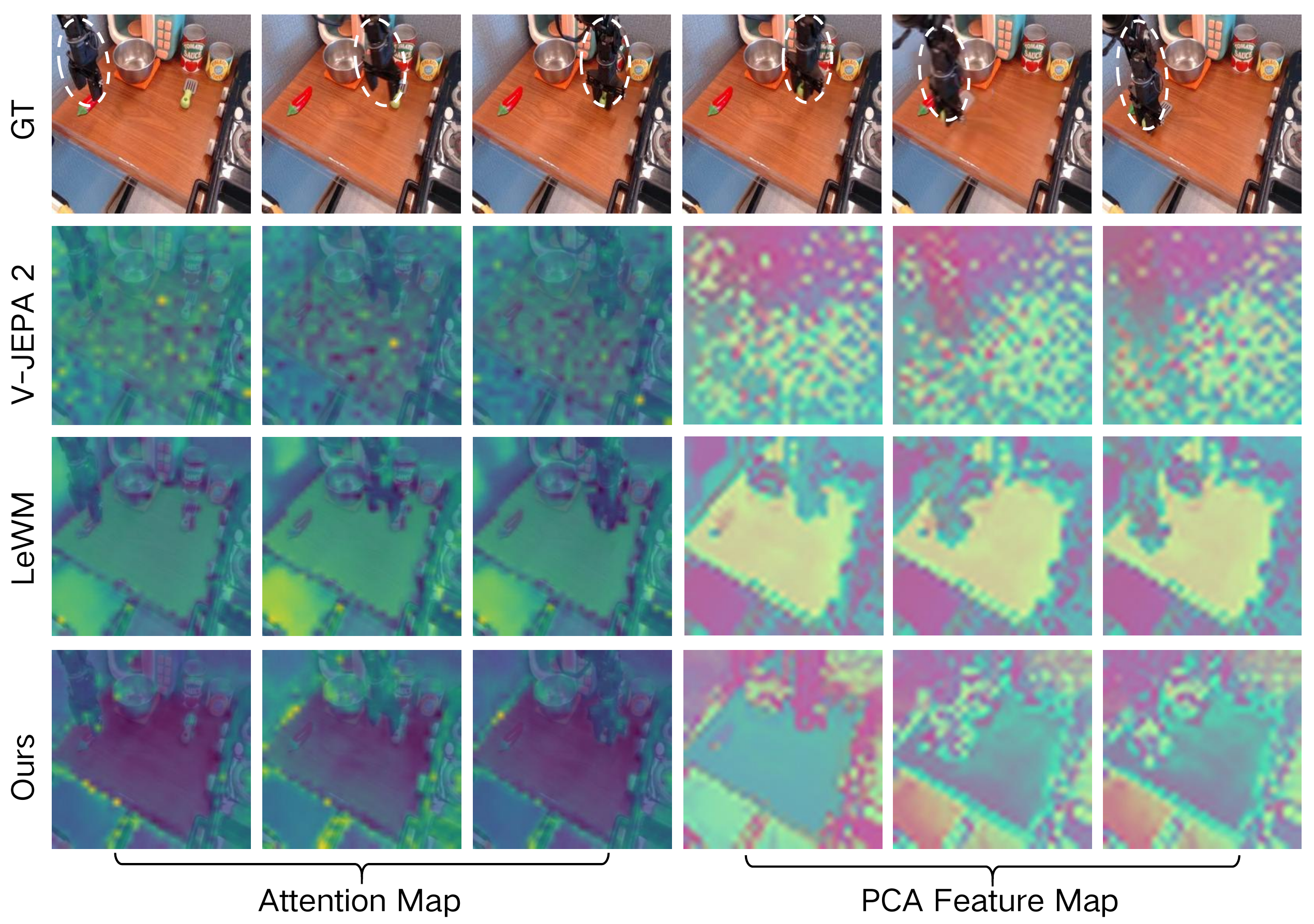}
 \caption{\textbf{Spatial feature structure.}
 Final-layer attention (left) and PCA of patch embeddings (right) for
 V-JEPA~2, LeWM, and our control-aligned encoder trained with the Dynamic CLS loss (Ours).
 LeWM focuses on the manipulation region, while Ours shows more concentrated
 attention along robot and object boundaries than LeWM.
 PCA is fitted separately on the same frames;
 colors are comparable only within each encoder.}
 \label{fig:app-attention}
\end{figure}

Figure~\ref{fig:app-attention} shows final-layer attention and PCA of
patch embeddings. V-JEPA~2 distributes attention across the scene and
exhibits fine-scale PCA variation. LeWM concentrates on the manipulation
region, with PCA maps that distinguish the robot, bowl, and object from
much of the tabletop. Compared with LeWM, our control-aligned encoder,
trained with the Dynamic CLS loss, shows more concentrated attention
along the edges of the robot and manipulated objects, outlining their
boundaries more clearly while retaining this foreground--background
separation in its PCA maps. These maps indicate where
information is represented; they do not identify the pixels used by the
downstream policy.

\begin{figure}[htbp]
 \centering
 \includegraphics[width=0.95\linewidth]{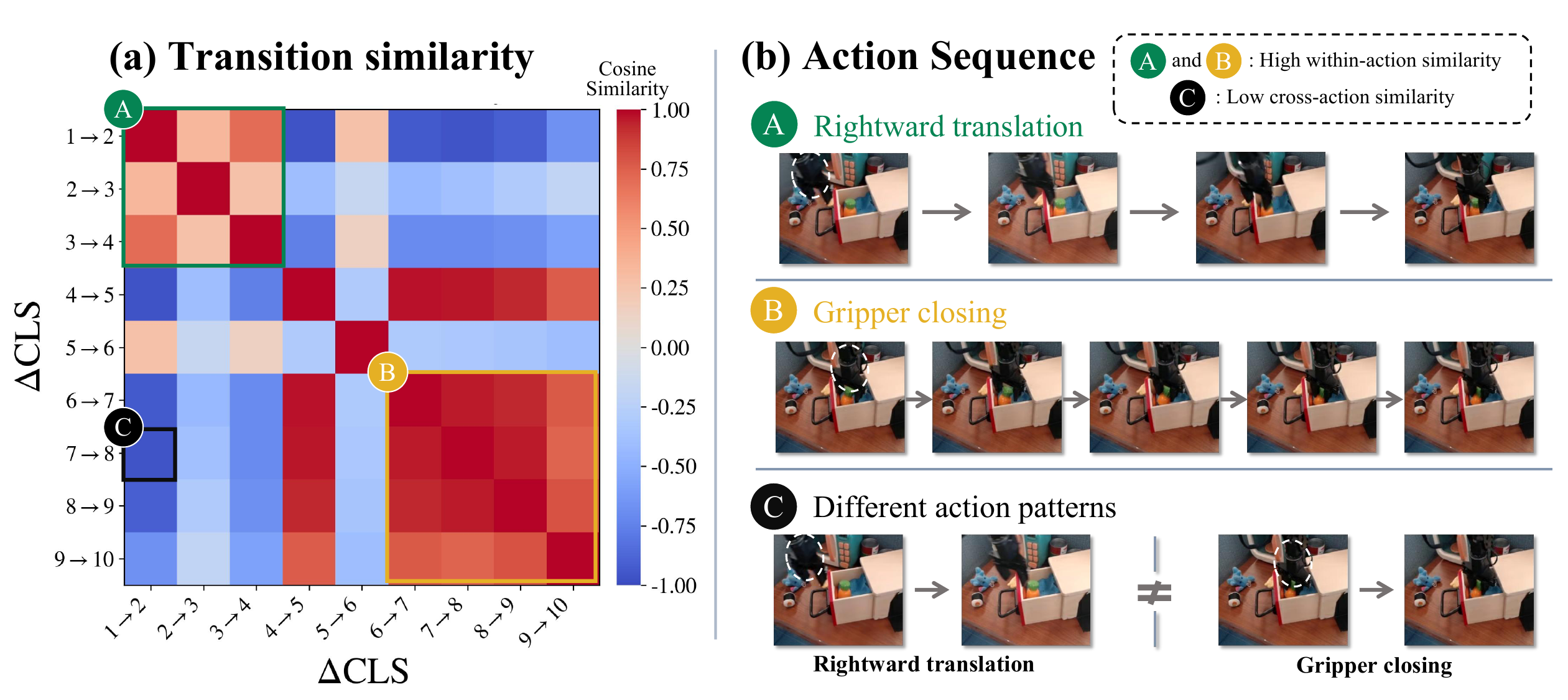}
 \caption{\textbf{CLS transition similarity across action patterns.}
 (a) Pairwise cosine similarity between CLS transition vectors
 $\Delta c_t=c_{t+1}-c_t$; warmer colors indicate higher similarity.
 The green block (A) covers rightward-translation transitions
 $1\!\to\!2$ through $3\!\to\!4$, and the yellow block (B) covers
 gripper-closing transitions $6\!\to\!7$ through $9\!\to\!10$.
 The black-outlined cell (C) contrasts $1\!\to\!2$ with $7\!\to\!8$,
 illustrating low similarity across the two action patterns.
 (b) Corresponding observation sequences and the cross-action comparison;
 dashed white circles highlight the gripper region.}
 \label{fig:app-cls-transition}
\end{figure}

\subsection{CLS Transition Similarity Across Action Patterns}
\label{app:cls-transition}

We next examine
whether changes in the global CLS token reflect the type of motion, using
the control-aligned encoder introduced in Method. For the ten observations
in the illustrated sequence, let $c_t=\Phi_\theta^{\mathrm{cls}}(o_t)$ and
$\Delta c_t=c_{t+1}-c_t$, following the notation of
Eq.~\ref{eq:method-dynamic-loss}. Figure~\ref{fig:app-cls-transition}
compares the nine transition vectors through
$S_{ij}=\operatorname{cos}(\Delta c_i,\Delta c_j)$, for
$i,j\in\{1,\ldots,9\}$. Each axis entry denotes an observation transition,
so the matrix measures similarity between representation changes rather
than between individual frame embeddings.

The off-diagonal entries within A and B show higher similarity among
transitions with the same action pattern: rightward translation in A and
gripper closing in B. In contrast, C compares a translation transition
with a closing transition and exhibits low similarity. In this example,
the global-token changes therefore distinguish the two motion patterns
while remaining more consistent within each pattern. This observation
is consistent with the motivation of the Dynamic CLS loss to encode
motion-relevant changes in the global state. The visualization concerns
transition direction in representation space; it does not measure
transition magnitude or, by itself, isolate the effect of the loss.

\section{Training Details}
\label{app:training}

\subsection{Training Data}
\label{app:training-data}

\paragraph{Bridge/Fractal.}
SimplerEnv policies are trained on BridgeData V2 and Fractal in LeRobot
format (Table~\ref{tab:app-data-inventory}). The sampler first selects a
source with probability $1/2$, then an episode and timestep uniformly
within that source, so equal source weights are not uniform over all
frames. The two datasets keep their native rates (5\,Hz and 3\,Hz);
frame offsets are counted in dataset steps. The policy uses one RGB view, resized to
$224\times224$. Each example contains the current image and instruction,
a contiguous 16-step action chunk, and eight future images at stride 2
for latent alignment. Indices past the episode end are clamped to the
last frame. Future images are training targets only.

\begin{table}[!htbp]
 \centering
 \caption{\textbf{Bridge/Fractal data inventory.}
 Encoder pretraining windows use four observations at stride two and do not cross
 episode boundaries.}
 \label{tab:app-data-inventory}
 \small
 \begin{tabular}{@{}lrrrr@{}}
 \toprule
 Dataset & Episodes & Frames & Rate (Hz) & Pretraining windows \\
 \midrule
 BridgeData V2 & 53,192 & 1,893,026 & 5 & 1,520,923 \\
 Fractal & 87,212 & 3,786,400 & 3 & 3,178,323 \\
 \bottomrule
 \end{tabular}
\end{table}

\paragraph{RoboCasa-GR1.}
We train one multitask policy on the 24 GR1 Tabletop Tasks with equal
task-sampling weights; within a task, an episode and timestep are drawn
uniformly. The policy receives the ego-view RGB image at $224\times224$
and the task instruction, without proprioception. Actions are 16-step
chunks with 29 channels (two arms, two hands, and waist), min--max
normalized to $[-1,1]$. Latent alignment uses eight future images at
stride 2, as in Bridge/Fractal training.

\paragraph{Real-world.}
We initialize from the SimplerEnv policy and fine-tune on 60
leader--follower demonstrations of ``Put the red cube into the blue
bowl'' (20 at each of three cube positions) on the right PiPER arm of an
AgileX Cobot Magic platform. Acquisition records top-view and
right-wrist RGB at $640\times480$ and 30\,Hz; the training adapter uses
the top view at $224\times224$ and a 15-fps export. HDF5 poses are
converted from millimeters/degrees to meters/radians. Translational
actions are consecutive position differences; rotational actions are
base-frame XYZ Euler increments from $R_{t+1}R_t^{-1}$; the gripper
encodes the next open/closed state in $\{0,1\}$. Trailing stationary
frames are trimmed. The exported 8D state is stored but not consumed by
the policy.

\subsection{Policy Optimization}
\label{app:training-hyperparameters}

Table~\ref{tab:app-training-hyperparameters} lists the three recipes.
All use Qwen3-VL-4B-Instruct with a GR00T flow-matching decoder,
AdamW $(\beta_1,\beta_2)=(0.9,0.95)$, $\epsilon=10^{-8}$, weight decay
$10^{-8}$, gradient clipping 1.0, cosine decay to $5\times10^{-7}$,
random seed 42, BF16, and DeepSpeed ZeRO-2. The backbone, predictive
branch, action decoder, and fusion modules are trainable; the pretrained control-aligned
teacher is frozen. The predictive branch is initialized by copying the
corresponding Qwen text-layer weights. Each action chunk is trained with
eight flow-matching noise/time samples; inference uses four Euler steps.
Patch fusion uses eight attention heads and residual gain $\gamma=0.2$.
There are $K=8$ predictive queries, matching the eight future targets.

Alignment follows Eq.~\ref{eq:method-align-loss} with
$\lambda_{\mathrm{sl1}}=\delta=0.1$. The coefficient $\alpha(s)$ in
$\mathcal L=\mathcal L_{\mathrm{FM}}+\alpha(s)\mathcal L_{\mathrm{align}}$
warms up linearly over $W$ steps, stays at 1 for $P$ steps, then
half-cosine decays over $D$ steps to 0.05 (Table~\ref{tab:app-training-hyperparameters}).
For Bridge/Fractal and RoboCasa, decay starts at 35{,}000 updates and
does not quite reach 0.05 within the 100{,}000-step budget.
Learning-rate warmup and alignment warmup are separate schedules.

\begin{table}[t]
 \centering
 \caption{\textbf{Policy optimization.}
 Shared optimizer, precision, and backbone settings are described in the
 text.}
 \label{tab:app-training-hyperparameters}
 \small
 \setlength{\tabcolsep}{4pt}
 \renewcommand{\arraystretch}{1.12}
 \begin{tabular}{@{}lccc@{}}
 \toprule
 Parameter & Bridge/Fractal & RoboCasa-GR1 & PiPER fine-tuning \\
 \midrule
 Update budget & 100,000 & 100,000 & 50,000 \\
 Devices $\times$ batch per device & $8\times16$ & $8\times8$ & $8\times16$ \\
 Global batch size & 128 & 64 & 128 \\
 Backbone learning rate & $10^{-5}$ & $10^{-5}$ & $10^{-4}$ (LoRA) \\
 Predictive-branch learning rate & $10^{-4}$ & $10^{-4}$ & $10^{-4}$ (LoRA) \\
 Action-decoder learning rate & $10^{-4}$ & $10^{-4}$ & $10^{-5}$ \\
 Other trainable modules' rate & $10^{-5}$ & $3\times10^{-5}$ & $5\times10^{-6}$ \\
 Learning-rate warm-up & 5,000 & 5,000 & 500 \\
 Checkpoint interval & 5,000 & 10,000 & 5,000 \\
 Action dimension / chunk length & $7/16$ & $29/16$ & $7/16$ \\
 Alignment warm-up / plateau / decay & 5k / 30k / 70k & 5k / 30k / 70k & 0.5k / 10k / 20k \\
 \bottomrule
 \end{tabular}
\end{table}

PiPER fine-tuning starts from a Bridge/Fractal checkpoint at 95{,}000
updates and retains $\mathcal L_{\mathrm{align}}$. LoRA (rank 32,
$\alpha=64$, dropout 0.05) is applied to attention and MLP projections
in the backbone and predictive branch, with those base weights frozen.
The action decoder and remaining fusion, projection, and query
parameters are updated directly.

\subsection{Control-Aligned Encoder Pretraining}
\label{app:encoder-training}

The control-aligned encoder is trained on Bridge/Fractal windows of four frames at offsets
$0,2,4,6$ (three context observations and one prediction target). Each
stride-two transition is conditioned on the two intervening 7D actions,
concatenated to 14D. Unlike the equal-source policy sampler, windows are
shuffled, so source mass follows Table~\ref{tab:app-data-inventory}
(about 32\% Bridge / 68\% Fractal). A seeded 90/10 split is applied at
the window level, so an episode can contribute to both subsets.

The encoder is ViT-Tiny ($224\times224$, patch 14, 768D hidden) with a
192D latent projector. The action-conditioned predictor has six layers,
16 heads, and dropout 0.1. Training uses 8 devices, batch 256 per device
(global 2{,}048), AdamW at $5\times10^{-5}$, weight decay $10^{-3}$,
gradient clipping 1.0, and BF16. SIGReg has weight 0.2 (1{,}024
projections, 17 knots). Dynamic CLS has weight 0.2 and temperature
$\tau=0.5$; its patch targets come from a frozen copy of the source
encoder. The policy uses the epoch-36 export of this run.

\section{Evaluation Details}
\label{app:evaluation}

\subsection{SimplerEnv}
\label{app:eval-simpler}

We evaluate the official WidowX Bridge suite: spoon-on-towel,
carrot-on-plate, green-on-yellow stacking, and eggplant-in-basket.
Object poses are randomized in episode mode over indices $0$--$23$
(24 episodes per task). We repeat the suite over four independent runs
(96 episodes per task). Control is 5\,Hz, simulation is 500\,Hz, and
each episode is limited to 120 control steps. Robot base orientation is
held at the identity. The first three tasks use the Bridge table with
base $(x,y)=(0.147,0.028)$\,m; eggplant uses the sink-camera setup with
base $(0.127,0.06)$\,m. Success is the binary outcome from the official
evaluator. We report the mean over the four runs and the unweighted mean
across tasks (Table~\ref{tab:simplerenv}). The \textit{Juno-TTT} row uses
the same launcher after test-time training. Policy inference is served
in BF16 over WebSocket to the simulator client.

\subsection{RoboCasa-GR1}
\label{app:eval-robocasa}

The multitask policy is evaluated on all 24 GR1 Tabletop Tasks in
Table~\ref{tab:robocasa-full} (GR1 arms and waist with Fourier hands), with
50 episodes per task, a 720-step horizon, and 12 executed actions from
each predicted chunk. An episode counts as success if the simulator
success flag is true at any step. We report per-task rates and their
unweighted mean. As in SimplerEnv, a BF16 policy server communicates
with the client over WebSocket.

Table~\ref{tab:robocasa-full} provides the full per-task breakdown
corresponding to the grouped results in Table~\ref{tab:robocasa}.
\begin{table*}[htbp]
  \centering
  \caption{Full per-task results on RoboCasa-GR1 Tabletop Tasks.
  A single multitask model is trained for all 24 tasks. Entries are success
  rates (\%); Average is the unweighted mean over all tasks. The best result
  in each row is shown in \textbf{bold}.}
  \label{tab:robocasa-full}
  \small
  \renewcommand{\arraystretch}{1.15}
  \setlength{\tabcolsep}{4.5pt}
  \begin{adjustbox}{max width=\textwidth}
  \begin{tabular}{@{}lcccccc@{}}
    \toprule
    \textbf{Task}
     & \textbf{GR00T-N1.6}
     & \textbf{Qwen3GR00T}
     & \textbf{Qwen3PI}
     & \textbf{Qwen3OFT}
     & \textbf{Qwen3FAST}
     & \textbf{\methodname{} (Ours)} \\
    \midrule
    PnPBottleToCabinetClose                & 51.5           & 46.0 & 26.0 & 30.0 & 38.0 & \textbf{80.0} \\
    PnPCanToDrawerClose                    & 13.0           & 80.0 & 62.0 & 76.0 & 44.0 & \textbf{86.0} \\
    PnPCupToDrawerClose                    & 8.5            & 54.0 & 42.0 & 44.0 & \textbf{56.0} & 48.0 \\
    PnPMilkToMicrowaveClose                & 14.0           & 48.0 & \textbf{50.0} & 44.0 & 44.0 & 44.0 \\
    PnPPotatoToMicrowaveClose              & 41.5           & 28.0 & \textbf{42.0} & 32.0 & 14.0 & 32.0 \\
    PnPWineToCabinetClose                  & 16.5           & 46.0 & 32.0 & 36.0 & 14.0 & \textbf{54.0} \\
    \addlinespace[4pt]
    PnPNovelFromCuttingboardToBasket       & 58.0           & 48.0 & 40.0 & 50.0 & 54.0 & \textbf{60.0} \\
    PnPNovelFromCuttingboardToCardboardbox & 46.5           & 40.0 & 46.0 & 40.0 & 42.0 & \textbf{58.0} \\
    PnPNovelFromCuttingboardToPan          & 68.5           & 68.0 & 60.0 & 70.0 & 58.0 & \textbf{78.0} \\
    PnPNovelFromCuttingboardToPot          & 65.0           & 52.0 & 40.0 & 54.0 & 58.0 & \textbf{68.0} \\
    PnPNovelFromCuttingboardToTieredbasket & 46.5           & 56.0 & 44.0 & 38.0 & 40.0 & \textbf{60.0} \\
    \addlinespace[4pt]
    PnPNovelFromPlacematToBasket           & 58.5           & 42.0 & 44.0 & 32.0 & 36.0 & \textbf{66.0} \\
    PnPNovelFromPlacematToBowl             & 57.5           & 44.0 & 52.0 & 58.0 & 38.0 & \textbf{64.0} \\
    PnPNovelFromPlacematToPlate            & 63.0           & 48.0 & 50.0 & 52.0 & 42.0 & \textbf{70.0} \\
    PnPNovelFromPlacematToTieredshelf      & \textbf{28.5} & 18.0 & 28.0 & 24.0 & 18.0 & 22.0 \\
    \addlinespace[4pt]
    PnPNovelFromPlateToBowl                & 57.0           & 60.0 & 52.0 & 60.0 & 52.0 & \textbf{62.0} \\
    PnPNovelFromPlateToCardboardbox        & 43.5           & 50.0 & 40.0 & 50.0 & 30.0 & \textbf{64.0} \\
    PnPNovelFromPlateToPan                 & 51.0           & 54.0 & 36.0 & 66.0 & 48.0 & \textbf{76.0} \\
    PnPNovelFromPlateToPlate               & \textbf{78.7} & 70.0 & 48.0 & 68.0 & 50.0 & 70.0 \\
    \addlinespace[4pt]
    PnPNovelFromTrayToCardboardbox         & 51.5           & 38.0 & 34.0 & 44.0 & 28.0 & \textbf{54.0} \\
    PnPNovelFromTrayToPlate                & \textbf{71.0} & 56.0 & 64.0 & 56.0 & 34.0 & 70.0 \\
    PnPNovelFromTrayToPot                  & \textbf{64.5} & 50.0 & 44.0 & 62.0 & 46.0 & 60.0 \\
    PnPNovelFromTrayToTieredbasket         & \textbf{57.0} & 36.0 & 50.0 & 54.0 & 36.0 & 48.0 \\
    PnPNovelFromTrayToTieredshelf          & 31.5           & 16.0 & 28.0 & 30.0 & 16.0 & \textbf{36.0} \\
    \midrule
    \textbf{Average}                       & 47.6           & 47.8 & 43.9 & 48.8 & 39.0 & \textbf{59.6} \\
    \bottomrule
  \end{tabular}
  \end{adjustbox}
  \vspace{-0.5em}
\end{table*}

\subsection{Real-world Evaluation}
\label{app:eval-real}

\begin{figure}[t]
    \centering
    \includegraphics[width=0.80\linewidth]{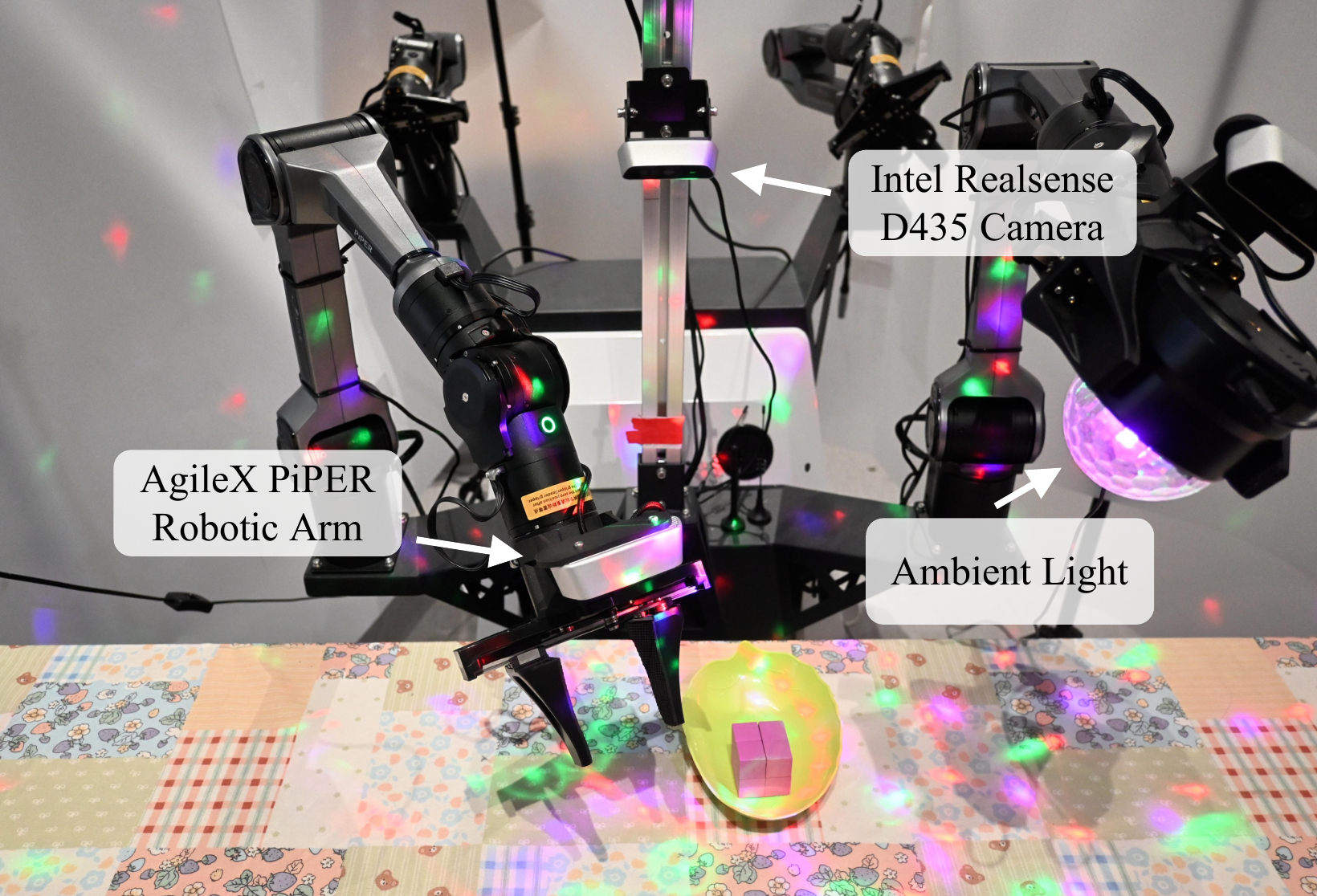}
    \caption{\textbf{Real-robot workstation with additional colored illumination.}
    This lighting is used for dynamic-lighting test-time training, not for
    the frozen-policy columns in Table~\ref{tab:real-world}.}
    \label{fig:real-ambient-setup}
\end{figure}

\paragraph{Platform and observations.}
\label{ttt:moresteup}
We use the right PiPER follower arm of an AgileX Cobot Magic
ALOHA-style platform (six revolute joints; parallel-gripper opening
$0$--$70$\,mm). The left arm and mobile base are unused. Inference runs
in BF16 on an NVIDIA DGX Spark; a ROS Noetic client on the robot PC
exchanges images, instructions, and action chunks over a wired
WebSocket. Top-view and right-wrist RGB are recorded with Intel
RealSense D435 cameras at $640\times480$ and 30\,Hz. The policy consumes
only the top view, resized to $224\times224$, as in fine-tuning.
Figure~\ref{fig:real-ambient-setup} shows the colored lamp used only in
the dynamic-lighting TTT setting (Fig.~\ref{fig:real-setup}c).

\paragraph{Control.}
The policy outputs a $16\times7$ delta end-effector chunk in the robot
base frame (meters, radians, binary gripper), denormalized with the
checkpoint's 1st/99th-percentile statistics. Deployment runs at 15\,Hz
and executes the full 16-step chunk ($\approx 1.07$\,s) before the next
observation and replan, unlike the 5\,Hz SimplerEnv protocol. Each chunk
is integrated from fresh end-effector feedback; inference is
synchronous and the previous Cartesian target is held while waiting.
Increments are converted by damped least-squares IK on the PiPER URDF.
Targets outside workspace or joint limits are rejected. If the IK
residual exceeds $0.01$\,m or $0.10$\,rad, that action is skipped.
Latched tracking faults require an operator reset and count as failures.

\paragraph{Protocol.}
We run 20 frozen-policy trials per model and condition
(Table~\ref{tab:real-world}). Every condition, including in-domain, uses
cube positions held out from the three demonstration placements.
In-domain keeps nominal appearance and objects. The remaining columns
are cumulative (Fig.~\ref{fig:real-setup}b): background replaces the
tablecloth and changes the cube color and the bowl's shape and color;
height additionally raises the work plane by $10$\,cm; object further
replaces the cube with a grape and updates the instruction. Success is
placing the designated object in the designated bowl without operator
intervention.

\section{Test-Time Training}
\label{app:ttt}

\subsection{Procedure}
\label{app:ttt-setup}

TTT is reported separately from frozen-policy evaluation because it uses
additional deployment interaction and optimization. A recorded transition
$(o_t,a_t,o_{t+1})$ is valid supervision for action-conditioned dynamics
whether or not the action completes the task; using $a_t$ as an imitation
target is not. Stage~A therefore adapts the world model on all
deployment transitions, including failures
(Eq.~\ref{eq:method-ttt-world}). Stage~B freezes this teacher and updates
the policy on successful data only, with joint alignment and
flow-matching (Eq.~\ref{eq:method-ttt-policy}). Stage~B updates the backbone and predictive branch through LoRA adapters while keeping their pretrained weights frozen. The action
decoder is fully trainable without LoRA, and the adapted teacher
remains frozen throughout policy re-alignment. Alignment weight is
held at 1 throughout Stage~B.

\subsection{SimplerEnv}
\label{app:ttt-sim}

SimplerEnv TTT uses rollouts of the frozen Bridge/Fractal policy as the
deployment stream. Stage~A trains all world-model parameters for 10
epochs with AdamW at $10^{-6}$, weight decay $10^{-3}$, batch
$8\times16$, BF16, SIGReg weight 0.2, and an episode-level 90/10 split.
Windows keep three context frames and one prediction target at stride 2.
Stage~B runs 400 steps with batch $8\times4$, DeepSpeed ZeRO-2,
a learning rate of $5\times10^{-7}$ for the backbone LoRA adapters
and $5\times10^{-6}$ for the other LoRA adapters and
the fully trainable action head,
10-step warmup, and cosine decay to $5\times10^{-8}$. Action chunks
remain length 16 with eight future frames at stride 2; inference uses
four Euler steps.

\subsection{Real Robot}
\label{app:ttt-real}

For the 15\,Hz dynamic-lighting condition, we collect 40 deployment
rollouts. Stage~A uses all valid transitions from these rollouts,
including those from failed executions. Stage~B uses only transitions
from verified successful executions.
RGB supervision is the top camera at $224\times224$. Control matches
Appendix~\ref{app:eval-real} (15\,Hz, 16-step chunks). Action
normalization is taken from PiPER fine-tuning and is not refit on the
deployment rollouts. The world model is warm-started from the Bridge/Fractal
frozen-CLS checkpoint; the policy is warm-started from the PiPER-adapted
checkpoint after 5{,}000 updates. Stage~A follows the SimplerEnv
world-model recipe for 10 epochs. Stage~B
follows the 400-step joint fine-tune above, with checkpoints every 50
steps. These updates are not applied in Table~\ref{tab:real-world}.

\section{Qualitative Results}
\label{app:qualitative}

We present paired rollouts in simulation and on the real robot to illustrate
policy behavior before and after the improvements studied in this work.
In all four figures, the upper row shows a failed execution and the lower
row shows a successful execution, with frames ordered from left to right.
Some pairs compare \methodname{} with the base VLA, and others compare
\methodname{} before and after test-time training. The lighting example is
the former; the Gaussian-noise example is the latter.

\subsection{Simulation Evaluation}
\label{app:qualitative-sim}

\begin{figure}[htbp]
    \centering
    \includegraphics[pagebox=mediabox,trim=253.2bp 249.6bp 23.1bp 72.8bp,clip,width=\linewidth]{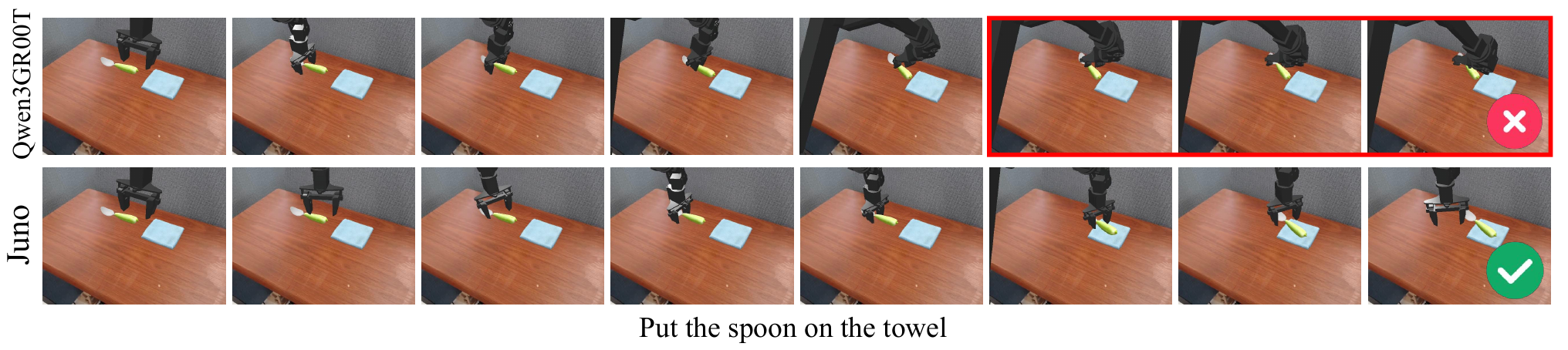}
    \caption{\textbf{Simulation: base VLA versus \methodname{}.}
    Spoon placement in SimplerEnv. The upper row shows an unsuccessful
    Qwen3GR00T rollout; the lower row shows a successful \methodname{}
    rollout that places the spoon on the towel. Frames progress from
    left to right.}
    \label{fig:qualitative-sim-base}
\end{figure}

\paragraph{Base VLA versus \methodname{}.}
Figure~\ref{fig:qualitative-sim-base} compares Qwen3GR00T and
\methodname{} on the SimplerEnv task ``Put the spoon on the towel.''
The baseline rollout in the upper row fails to place the spoon on the
towel. In contrast, \methodname{} in the lower row picks up the spoon,
transports it to the towel, and releases it on the target surface.
This pair illustrates how the performance difference reported in
Table~\ref{tab:simplerenv} can manifest in a complete manipulation sequence.

\begin{figure}[htbp]
    \centering
    \includegraphics[pagebox=mediabox,trim=231.8bp 60.4bp 44.4bp 262.0bp,clip,width=\linewidth]{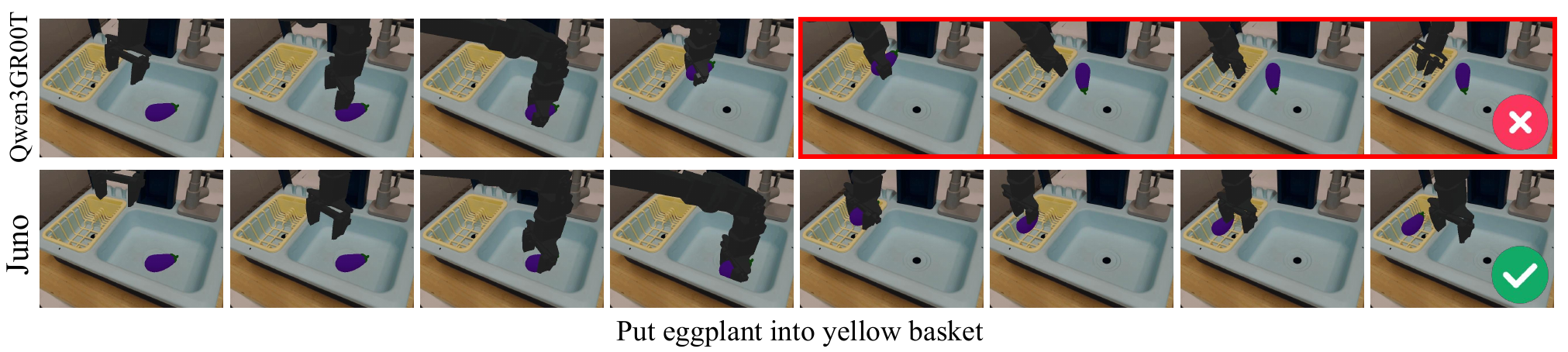}
    \caption{\textbf{Simulation: \methodname{} before and after TTT.}
    Eggplant placement in SimplerEnv. The upper row shows a failed
    \methodname{} execution before adaptation; the lower row shows
    successful placement into the yellow basket after TTT (Juno-TTT
    in the figure). Frames progress from left to right.}
    \label{fig:qualitative-sim-ttt}
\end{figure}

\paragraph{\methodname{} before and after TTT.}
Figure~\ref{fig:qualitative-sim-ttt} compares \methodname{} before and
after TTT on ``Put eggplant into yellow basket.'' In the upper row,
the policy lifts the eggplant but does not complete its transfer into
the basket. In the lower row, the adapted policy grasps the eggplant,
moves it from the sink to the yellow basket, and completes placement.
This is one rollout pair. Average success on this task stays essentially
unchanged after TTT ($96.9\%$ to $95.8\%$, Table~\ref{tab:simplerenv}).

\subsection{Real-world Evaluation}
\label{app:qualitative-real}

The real-robot examples show cube-to-bowl manipulation under Gaussian
observation noise and flickering colored lighting. As in simulation,
each figure pairs a failed rollout with a successful one, while the
model comparison differs between the two conditions.

\begin{figure}[htbp]
    \centering
    \includegraphics[pagebox=mediabox,trim=96.8bp 107.5bp 178.5bp 177.6bp,clip,width=\linewidth]{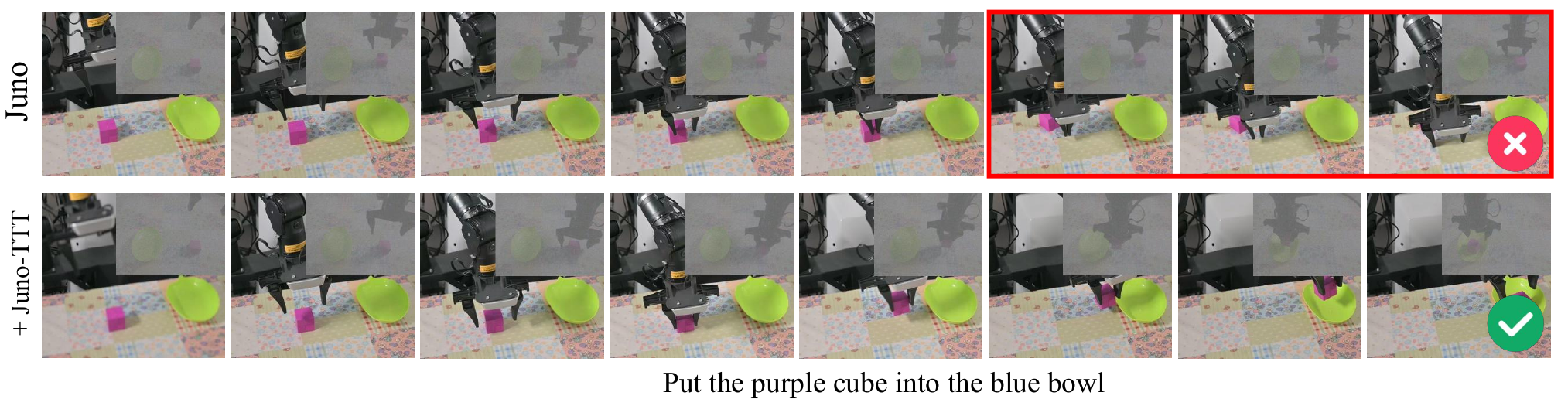}
    \caption{\textbf{Real robot: Gaussian observation noise.}
    The upper row shows a failed \methodname{} rollout before TTT;
    the lower row shows a successful rollout after TTT (Juno-TTT in
    the figure). Insets display the noise-corrupted observations.
    Frames progress from left to right, ending with the cube outside
    the bowl in the upper row and inside it in the lower row.}
    \label{fig:qualitative-real-gaussian}
\end{figure}

\paragraph{Gaussian observation noise: before and after TTT.}
Figure~\ref{fig:qualitative-real-gaussian} compares \methodname{}
before and after TTT with noise-corrupted observations, shown in the
frame insets. Before adaptation, the robot approaches the cube and
attempts to grasp it, but the cube remains outside the bowl in the
shown sequence. After TTT, the robot grasps and lifts the cube,
transports it over the bowl, and releases it inside. This pair
illustrates successful task completion after adaptation in the
presence of observation corruption.

\begin{figure}[htbp]
    \centering
    \includegraphics[pagebox=mediabox,trim=47.1bp 229.1bp 226.5bp 58.6bp,clip,width=\linewidth]{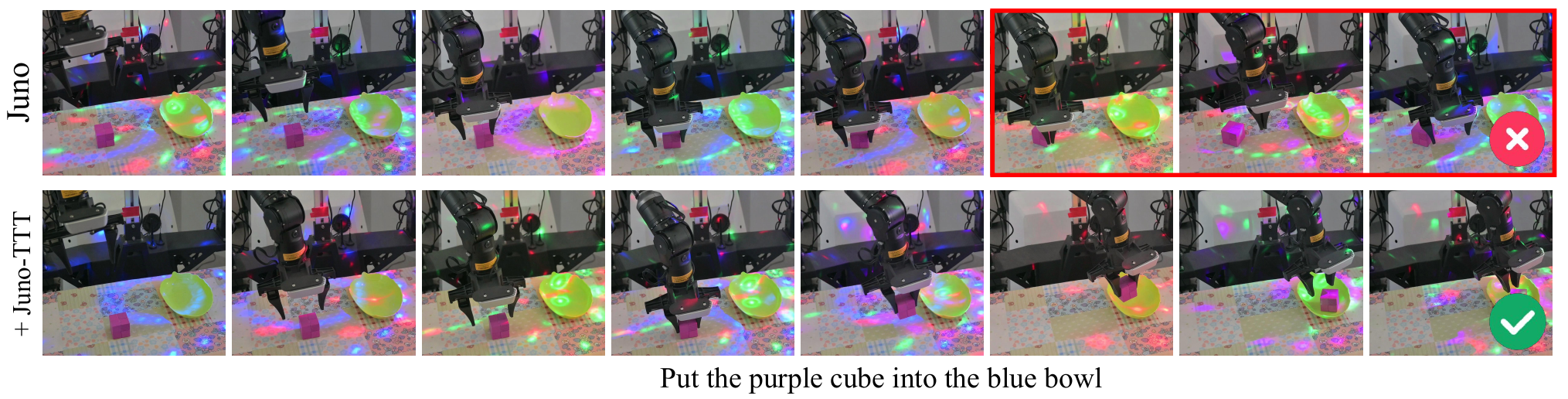}
    \caption{\textbf{Real robot: flickering colored lighting.}
    The upper row shows an unsuccessful Qwen3GR00T rollout; the lower
    row shows a successful \methodname{} rollout. Colored light
    patterns vary across the scene as the robot attempts to place
    the cube into the bowl. Frames progress from left to right.}
    \label{fig:qualitative-real-light}
\end{figure}

\paragraph{Flickering lighting: base VLA versus \methodname{}.}
Figure~\ref{fig:qualitative-real-light} compares Qwen3GR00T and
\methodname{} under flickering colored illumination. The changing
light patterns alter the appearance of the tabletop, robot, and
objects throughout the rollouts. In the upper row, the baseline
attempts to manipulate the cube but leaves it on the tabletop
outside the bowl. In the lower row, \methodname{} grasps the cube,
carries it to the bowl, and completes placement despite the changing
illumination. This comparison is between the frozen base policy and frozen
\methodname{}. The dynamic-lighting gain of \textit{Juno-TTT} is the
separate result in Table~\ref{tab:real-world-right} ($55\%$ to $70\%$).

Together, these examples illustrate differences in task completion
between the base VLA and \methodname{}, and between \methodname{}
before and after TTT. They complement the quantitative evaluation;
the selected rollouts alone do not establish success rates or isolate
the causes of individual failures.

\end{document}